\documentclass[10pt,twocolumn,letterpaper]{article}

\usepackage{cvpr}              % To produce the CAMERA-READY version
\usepackage{graphicx}
\usepackage{amsmath}
\usepackage{amssymb}
\usepackage{booktabs}
\usepackage{enumitem}

\usepackage[pagebackref,breaklinks,colorlinks]{hyperref}

\usepackage[capitalize]{cleveref}
\crefname{section}{Sec.}{Secs.}
\Crefname{section}{Section}{Sections}
\Crefname{table}{Table}{Tables}
\crefname{table}{Tab.}{Tabs.}

\usepackage[table]{xcolor}
\usepackage{bbding}
\usepackage{multirow}

\def\cvprPaperID{*****} % *** Enter the CVPR Paper ID here
\def\confName{CVPR}
\def\confYear{2023}

\begin{document}

%%%%%%%%% TITLE
\title{First Place Solution to the CVPR’2023 AQTC Challenge: \\ A Function-Interaction Centric Approach with Spatiotemporal Visual-Language Alignment}

% Tom Tongjia Chen1, Hongshan Yu1,2, Zhengeng Yang1, Ming Li3, Zechuan Li1, Jingwen Wang1, Wei Miao1, Wei Sun1, Chen Chen3
% 1College of Electrical and Information Engineering, Hunan University
% 2Quanzhou Institute of Industrial Design and Machine Intelligence Innovation, Hunan University
% 3Center for Research in Computer Vision, University of Central Florida

\author{Tom Tongjia Chen$^1$, Hongshan Yu$^{1,2}$,  Zhengeng Yang$^1$, Ming Li$^3$,\\ Zechuan Li$^1$, Jingwen Wang$^1$,
Wei Miao$^1$, Wei Sun$^1$, Chen Chen$^3$ \\
$^1$College of Electrical and Information Engineering, Hunan University \\
$^2$Quanzhou Institute of Industrial Design and Machine Intelligence Innovation, Hunan University\\ 
$^3$Center for Research in Computer Vision, University of Central Florida \\
{\tt\small $\{$tomchen, yuhongshan, yzg050215, lizechuan, wjw0375, weimiao$\}$@hnu.edu.cn; }\\
{\tt\small liming@knights.ucf.edu; david-sun@126.com; chen.chen@crcv.ucf.edu;}
}
\maketitle

%%%%%%%%% ABSTRACT
% \begin{abstract}
%    Egocentric AI assistant has garnered increasing attention from users. To address the core problem in egocentric assistant systems, a new task called Affordance-Centric Question-driven Task Completion (AQTC) has been defined. In this paper, we propose a \textbf{Parallel Function-Interaction Dual-Centric} approach to advance the AQTC task. We leverage well-aligned visual and textual encoders to bridge the gap between natural language instructions and visual operations. Additionally, we introduce a novel hand-object-interaction (HOI) aggregation module to enhance the model's ability to capture HOI contexts and strengthen the process of interaction grounding. Specifically, a pre-trained egocentric video-text model EgoVLP is employed to capture global temporal cues. The experimental results reveal that our method ranks first in the CVPR2023 AQTC Challenge, with a 78.7\% Recall@1.
%    Our code is available at \url{https://github.com/tomchen-ctj/CVPR23-LOVEU-AQTC}.
% \end{abstract}

% %%%%%%%%% ABSTRACT
% I rewrote the abstract, you can check for errors or need to modify
% I will finish writing the Introduction in the evening
\begin{abstract}
Affordance-Centric Question-driven Task Completion (AQTC) has been proposed to acquire knowledge from videos to furnish users with comprehensive and systematic instructions. However, existing methods have hitherto neglected the necessity of aligning spatiotemporal visual and linguistic signals, as well as the crucial interactional information between humans and objects. To tackle these limitations, we propose to combine large-scale pre-trained vision-language and video-language models, which serve to contribute stable and reliable multimodal data and facilitate effective spatiotemporal visual-textual alignment. Additionally, a novel hand-object-interaction (HOI) aggregation module is proposed which aids in capturing human-object interaction information, thereby further augmenting the capacity to understand the presented scenario. 
Our method achieved first place in the CVPR'2023 AQTC Challenge\footnote{https://sites.google.com/view/loveucvpr23/track3}, with a Recall@1 score of 78.7\%. The code is available at \url{https://github.com/tomchen-ctj/CVPR23-LOVEU-AQTC}.
\end{abstract}

\begin{figure}[t]
    \centering
    \includegraphics[width=1.0\linewidth]{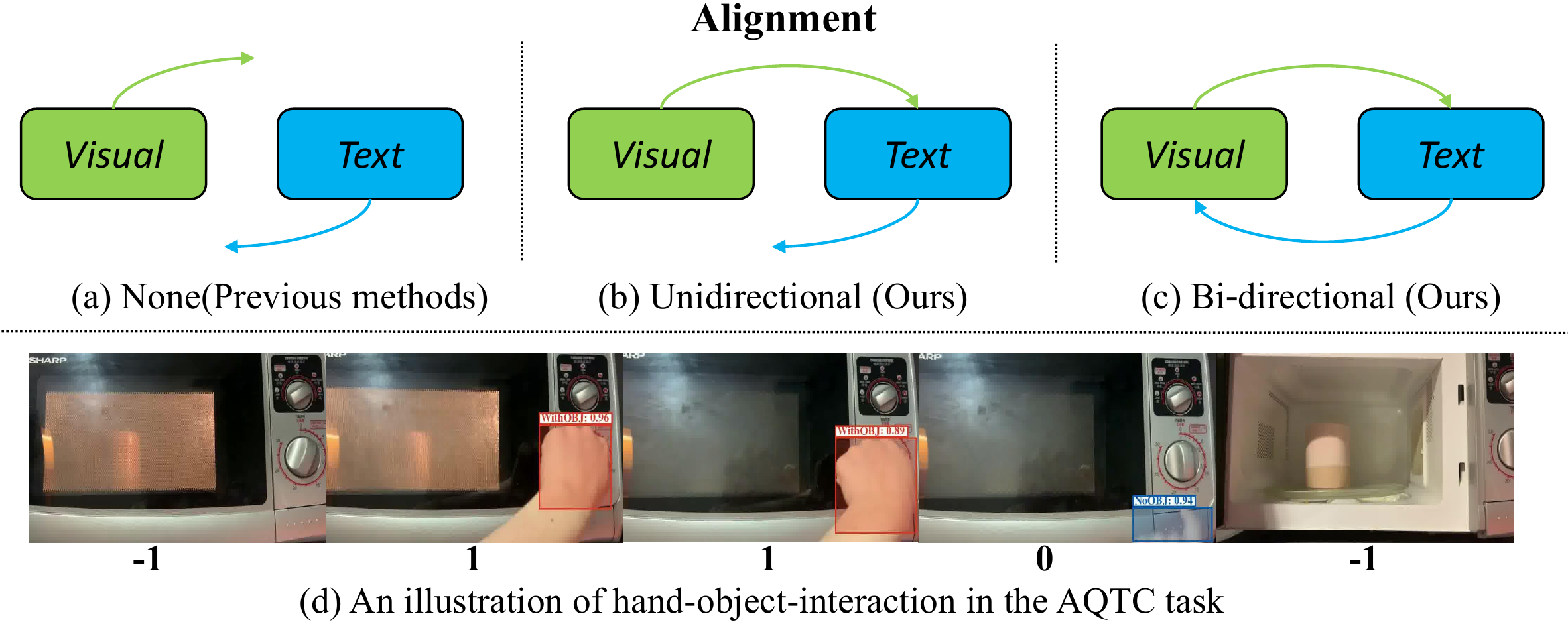}
    \caption{(a)-(c) Illustration of the various alignments between visual and textual encoders. (d) An illustration of hand-object interaction detection in AQTC task. Here, -1 denotes the absence of a hand, 0 denotes no human-object interaction, and 1 denotes the presence of interaction.}
    \label{fig:fig1}
\end{figure}  
%%%%%%%%% BODY TEXT
\section{Introduction}
\label{sec:intro}
% egocentic AI assistant
% AssistQ benchmark, difficulties compared to conventional VQA #1 multi-modal #2 egocentric #3 step-by-step
% Motivations: 1. alignment 2. temporal pooling 3. video model
% We rank 1 :)

The Affordance-Centric Question-driven Task Completion (AQTC) approach is designed to address a series of questions posed by users through a sequence of actions. While similar to traditional Visual Question Answering (VQA) methods, AQTC not only necessitates a clear comprehension of static images but also requires an understanding of changes in human and object interactions across space and time, which is a task that demands rigorous multimodal understanding and alignment capabilities. Moreover, the task-oriented nature of the questions in AQTC calls for multi-modal and multi-step answers, proving more challenging and akin to the intelligent assistant setting in comparison to VQA.

\begin{figure*}[t]
  \centering
  \includegraphics[width=1.0\textwidth]{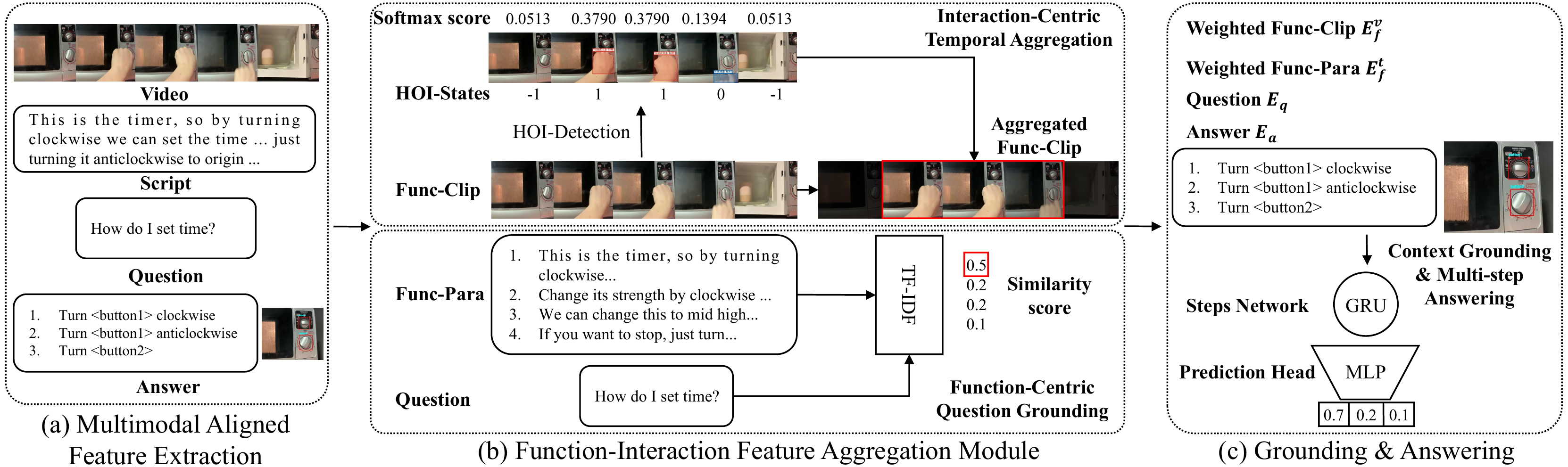}
  \caption{The pipeline of our Function-Interaction Centric approach. (a) Well-aligned multimodal encoders are utilized for feature extraction. (b) The script and video are first segmented into function-clips and function-paras to enable a fine-grained analysis. Then, the Interaction-Centric Temporal Aggregation module and Function-Centric Question Grounding module are employed to enhance the understanding of the interaction context and establish the relationship between questions and functions, respectively. (c) The weighted representations of function-paras and function-clips are combined with the question-and-answer embeddings to facilitate multiple-source context grounding. Next, a GRU followed by an MLP is employed to leverage historical steps and predict the final scores of each answer, taking into account the accumulated step-by-step contextual information.} 
  % Please briefly explain the pipeline here.
    \label{fig:2}
\end{figure*}

However, existing methods primarily rely on models pre-trained on a single modality for feature extraction, \emph{e.g.}, ViT\cite{vit}, XL-Net\cite{xlnet}, and BERT\cite{devlin2018bert}. In addition, they typically obtain video-level features through temporal average pooling and subsequently perform context grounding using techniques such as cross-attention\cite{assistq} or statistic-based question-function grounding\cite{wu2022winning}. As depicted in Figure \ref{fig:fig1}(a), these naive methods do not consider the spatiotemporal alignment of vision and text, which is crucial for multi-source context grounding. Also as shown in Figure \ref{fig:fig1}(d), in certain sections of the video where there is no interaction between the hand and object, the frames are just illustrations of the object. Pooling the frame-level features in these instances may diffuse the model's attention toward the crucial information provided by human-object interaction.   % Fig.1: motivation (figure and description)

% fig1 b, c is involved in the description of our methods
To address the aforementioned issues, we propose the utilization of visual-language pre-trained models~\cite{EGOVLP,BLIP,CLIP} to extract well-aligned  features as illustrated in Figure \ref{fig:fig1}(b)(c), along with an egocentric-video-language pre-trained model for improved comprehension of long-form temporal cues.
Additionally, we introduce the Interaction-Centric Temporal Aggregation module. It makes up for the defect that the previous Function-Centric method only focused on question grounding while ignoring the interaction between humans and objects, which contributes to extracting the contextual information pertaining to interactions within the extensive instructional videos. % Fig.2: Our Methods (Description)
 
The contributions of this work are summarized in four aspects:

\begin{itemize}[leftmargin=*]
\item We point out that existing methods are limited by the spatiotemporal visual-language misalignment and the lack of human-object interaction information.

\item We demonstrate that aligning the spatiotemporal visual-language information can achieve remarkable improvements, by simply combining visual-language and video-language foundation models.

\item We extend the Function-centric approach by introducing the novel Interaction-Centric Temporal Aggregation module for better task-scene understanding.

\item We achieve a 78.7\% Recall@1 score on the 2023 AQTC testing set, which takes the championship on the CVPR’2023 AQTC Challenge.
\end{itemize}

\begin{table*}[t]
  \centering
  \begin{tabular}{c|ccc|c|c|cc}
    \toprule
    Method      & Video Encoder    & Image Encoder  & Text Encoder & HOI          &   Alignment    & R@1     & R@3   \\
    \midrule
  Baseline\cite{wu2022winning}&ViT-L& ViT-L         &  XL-Net      & \XSolidBrush &     None       & 63.9    & 89.5  \\
    \midrule
  BLIP-Local    &  BLIP-ViT         & BLIP-ViT      &  BLIP-T      & \XSolidBrush & Bi-directional & 67.5    & 88.2  \\
  CLIP-Local    &  CLIP-ViT         & CLIP-ViT      &  XL-Net      & \Checkmark   & Unidirectional & 67.9    & 88.9  \\
  EgoVLP-Local  &  EgoVLP-V*        & EgoVLP-V*     &  EgoVLP-T    & \Checkmark   & Bi-directional & 74.1    & 90.2  \\
  EgoVLP-Global &  EgoVLP-V         & CLIP-ViT      &  EgoVLP-T    & \Checkmark   & Bi-directional & 74.8    & 91.5  \\
    \midrule
   Linear Ensemble & -              &  -            & -            & -            & -      & \textbf{78.7}   & \textbf{93.4}  \\
    \bottomrule
  \end{tabular}
  \caption{Results of different models and the impact of model ensemble. Here * indicates the extraction of image features using a video model, which involves padding along the temporal dimension. HOI denotes our Interaction-Centric temporal aggregation module.}
  \label{tab:1}
  \vspace{-5mm}
\end{table*}

%-------------------------------------------------------------------------
\section{Methodology}
\label{sec:method}
% fig1 teaser: motivation & intuition
% fig2 pipeline
% fig3 HOI aggregation (May integrate into fig2)
In this section, we present the technical details of the proposed Function-Interaction Centric framework. In the following subsections, we introduce the Multimodal Aligned Feature Extraction in Sec.~\ref{feature}, Function-Interaction Feature Aggregation in Sec.~\ref{aggregation}, and grounding and answering in Sec.~\ref{output}. The pipeline of our approach is illustrated in Figure \ref{fig:2}.

\begin{figure}[t]
    \centering
    \includegraphics[width=0.8\linewidth]{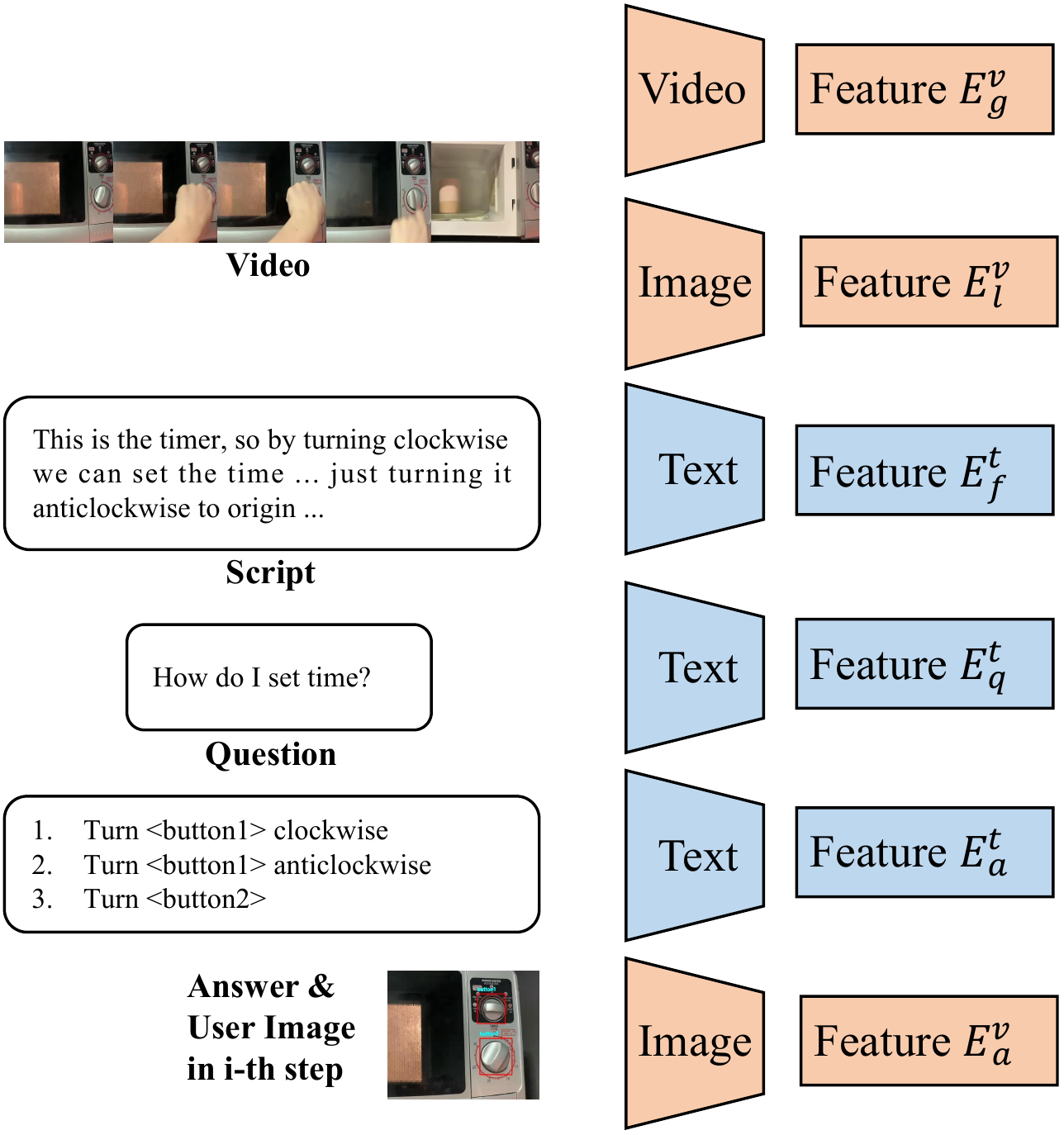}
    \caption{Illustration of the Multimodal Aligned Feature Extraction. We ensure alignment between the text and visual encoders we utilize. Specifically, for the global variant, we employ a video model to encode the long-form interaction contexts, while for the local variant, we utilize an image model.}
    \label{fig:fig3}
    \vspace{-5mm}
\end{figure}  
\subsection{Multimodal Aligned Feature Extraction} \label{feature}

The AQTC benchmark consists of multiple source data. We adopt the strategy employed in the baseline approach \cite{wu2022winning} to segment the script $S$ into textual function-paragraphs $f^t$ based on a pre-defined schema, and then divide the corresponding video $V$ into the visual function-clips $f^v$ via the aligned script timestamp. By implementing this approach, we partition the instructional video and script into a set of functions $\{f_1, f_2, \dots, f_n\}$. Each function denoted as $f$, not only comprises a textual description $f^t$ but also encompasses visual guidance $f^v$.

We observed that the alignment between text and visual encoders has a substantial impact on the final results. Hence, we employed well-aligned visual-language pre-trained models for feature extraction. The effects of feature alignment will be thoroughly discussed in Section \ref{ablation}.

We decode the function-clips at a frame rate of 1 and extract both local and global clip features using two model variants. In the local variant, we utilize an image encoder to extract frame-level features for each function-clip, denoted as $E_l^v=\{[E_1]_l^v, [E_2]_l^v, \dots, [E_n]_l^v\}$. On the other hand, the global variant extracts function-level features $E_g^v=\{[E_1]_g^v, [E_2]_g^v, \dots, [E_n]_g^v\}$ using a video model, as illustrated in Figure \ref{fig:fig3}.

For the global variant, we employ the video encoder from EgoVLP\cite{EGOVLP}, which is pre-trained on the large-scale egocentric dataset Ego4D\cite{ego4d}. This choice of encoder equips the model with the capability to capture long-form hand-object interactions, which is crucial for the AQTC task.

To ensure domain alignment, we employ the same encoder used for videos to encode button images into $E_a^v$. In particular, when utilizing the video model, we pad the button image along the temporal dimension. Similarly, we encode the function-paragraph, question, and candidate answers into $E_f^t$, $E_q^t$ and $E_a^t$ via the same text encoder. This ensures that the textual components are aligned in the feature space.

\subsection{Function-Interaction Feature Aggregation} \label{aggregation}
% fig2 for HOI aggregation
We now turn to explain the technical details of the Function-Interaction Feature Aggregation.

\textbf{Function-Centric Question Grounding.} 
Taking into account the relatively low semantic similarity between the corresponding functions and the given questions. Following\cite{wu2022winning}, we propose to employ a statistic-based TF-IDF\cite{ramos2003using} model to calculate the similarity score between the specific question $Q$ with the function set $\{f_1, f_2, \dots, f_n\}$. This choice has demonstrated superior effectiveness in small-scale data settings compared to the utilization of cross-attention.

\textbf{Interaction-Centric Temporal Aggregation.}
To better capture the hand-object-interaction information in long-form videos, we utilize an off-the-shelf HOI detector HI-RCNN\cite{chen2023affordance} which provides the states of interaction. The pipeline of the Interaction-Centric aggregation module is depicted in Figure \ref{fig:2}. % add Fig2.

First, we apply HI-RCNN to process the visual function-clips $f^v$ and generate HOI states $S^v$ at the frame level. Subsequently, we compute a softmax score for each function-clip. Finally, we perform a weighted combination within the current local clip to obtain the aggregated video feature $E_f^v$.

\subsection{Context Grounding \& Multi-step Answering} \label{output}
% simply MLP and gru
The multi-step QA task can be effectively formulated as a classification problem, given the available number of steps and candidate answers. Our goal is to accurately predict the correct action for each step, along with the corresponding button, based on the historical steps provided.

In line with the baseline approach\cite{wu2022winning}, we employ a MLP to perform context grounding. To leverage the information from the historical steps and make predictions, we utilize a GRU\cite{chung2014empirical} along with a two-layer MLP, followed by softmax activation, as the prediction head. This architecture enables us to effectively capture the step-by-step dependencies and predict the final score for each answer.
\begin{table}[!h]
  \centering
  \begin{tabular}{cc|cc}
    \toprule
  Visual Encoder &   Text Encoder   & R@1             & R@3            \\
    \midrule
  ViT            &   XL-Net         &         63.9    &         89.5   \\
  CLIP-ViT       &   XL-Net         &         66.6    & \textbf{90.8}  \\
  BLIP           &   BLIP           & \textbf{67.5}   &         88.2   \\
    \bottomrule
  \end{tabular}
  \caption{Ablation study on the alignment of pre-trained encoders. All the results are evaluated on the test set.}
  \label{tab:2}
  \vspace{-4mm}
\end{table}

\begin{equation}
    \hat{y_i} = HEAD(GRU(MLP([E_f, E_q^t, E_a]))),  \\
\end{equation}
where $E_f =\{E_f^t, E_f^v\}$ is the weighted function set, $E_q^t$ denotes the embeddings of question and $E_a=\{E_a^t, E_a^v\}$ denotes embeddings of the candidate answers. We use cross-entropy (CE) with softmax activation to calculate the loss.
\begin{equation}
    L = \sum\limits_{i=1}^{N} -y_i \log \hat{y_i}.  \\
\end{equation}

%------------------------------------------------------------------------
\section{Experiments}
\label{sec:exps}

\subsection{Experimental Settings}
% models (CLIP, BLIP, EgoVLP), benchmark 2023 train&test, implementation details
The 2023 AQTC challenge expanded the testing set by adding 100 additional QA pairs compared to the previous year. Our model was trained on the training set using the 2022 training and testing data. The evaluation metrics used, Recall@1 and Recall@3, remained consistent.

All experiments were conducted on a single NVIDIA GeForce RTX 3090 GPU. To ensure a fair comparison, we trained our model using the same parameter settings, including the Adam optimizer\cite{adam} with a learning rate of $1 \times 10^{-4} $ and 100 epochs. We report the evaluation metrics obtained at the epoch with the best performance on the testing set.

\subsection{Experimental Results}
% several variants, ensemble
Our best performance in the challenge was achieved by combining the classification scores of all models, which can be seen as a parallel model ensemble. As discussed in Sec. \ref{feature}, we employ various well-aligned feature extractors to ensure the utilization of reliable multimodal information, including CLIP, BLIP, and  EgoVLP\cite{CLIP, BLIP, EGOVLP}. The ensemble result achieved approximately a $4\%$ higher than the best-performing single model, as shown in Table \ref{tab:1}.

\subsection{Ablation Study} \label{ablation}
% HOI module, alignment

% \begin{table}[!h]
%   \centering
%   \begin{tabular}{c|cc|cc}
%     \toprule
%    Method                        &  Visual Encoder &   Text Encoder   & R@1             & R@3            \\
%     \midrule
%    Baseline\cite{wu2022winning}  &  ViT            &   XL-Net         &         63.9    &         89.5   \\
%    Replace visual side           &  CLIP-ViT       &   XL-Net         &         66.6    & \textbf{90.8}  \\
%    Replace both side             &  BLIP           &   BLIP           & \textbf{67.5}   &         88.2   \\
%     \bottomrule
%   \end{tabular}
%   \caption{The effect of model ensemble.}
%   \label{tab:3}
% \end{table}

\begin{table}[!h]
  \centering
  \begin{tabular}{cc|c|cc}
    \toprule
    Image Encoder    & Text Encoder   & HOI             & R@1                & R@3 \\
    \midrule
     EgoVLP-V        & EgoVLP-T       &  \XSolidBrush   &        72.1        & \textbf{91.8}  \\
     EgoVLP-V        & EgoVLP-T       &  \Checkmark     &\textbf{74.1}       &         90.2\\
    \midrule
    CLIP-ViT         & XL-Net         &  \XSolidBrush   &        66.6        & \textbf{90.8}  \\
    CLIP-ViT         & XL-Net         &  \Checkmark     &\textbf{67.9}       &         88.9 \\
    \bottomrule
  \end{tabular}
  \caption{Ablations on the effect of the interaction aggregation. HOI denotes Interaction-Centric temporal aggregation module.}
  \label{tab:3}
  \vspace{-4mm}
\end{table}

We conducted an ablation study on the alignment of visual and textual encoders, as presented in Table \ref{tab:2}. Initially, the baseline model employed ViT\cite{vit} as the visual encoder and XL-Net\cite{xlnet} as the textual encoder, without any alignment between them. To improve the alignment, we replaced the visual encoder with CLIP pre-trained ViT\cite{CLIP}, resulting in a unidirectional alignment that yielded a performance gain of $3\%$. Moreover, by replacing both sides of the model with BLIP\cite{BLIP}, which provides bi-directional alignment, we observed a further improvement of $4\%$ in Recall@1. 

We also conducted an ablation study on the Interaction-Centric aggregation module, as presented in Table \ref{tab:3}. The results demonstrate that our Interaction aggregation module significantly enhances Recall@1 by $1.3\%$ and $2\%$ for CLIP\cite{CLIP} and EgoVLP\cite{EGOVLP}, respectively. These findings provide strong evidence that our Interaction aggregation module plays a crucial role in improving affordance grounding.

\subsection{Visualization}
\begin{figure}[t]
    \centering
    \includegraphics[width=1\linewidth]{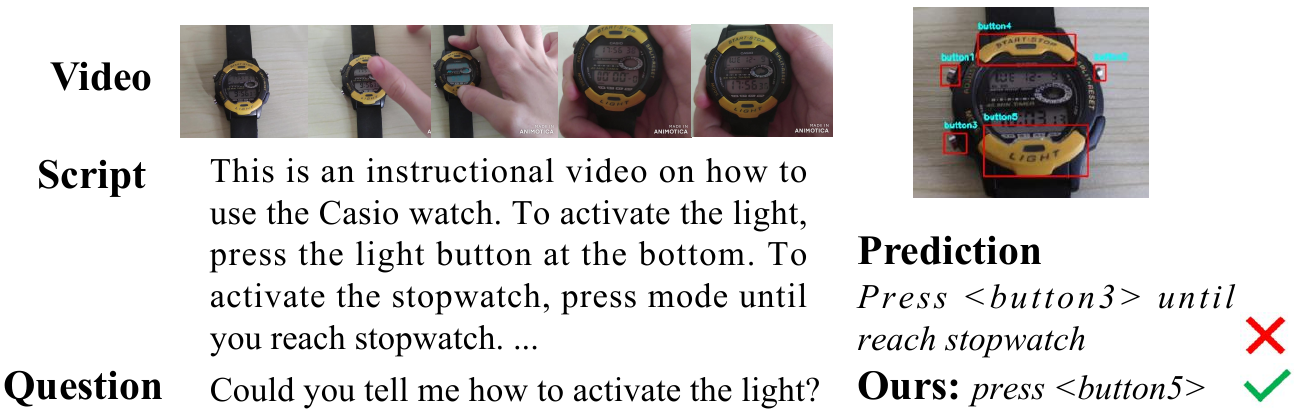}
    \caption{Qualitative result of our method.}
    \label{fig:fig4}
    \vspace{-4mm}
\end{figure}

Figure \ref{fig:fig4} visually represents the qualitative results of our method. By incorporating improved interaction context understanding and feature alignment, our approach demonstrates correct predictions compared to the baseline method.

% \begin{table}[!h]
%   \centering
%   \begin{tabular}{c|cc|c|cc}
%     \toprule
%     Method      & Image Encoder    & Text Encoder   & HOI             & R@1                & R@3 \\
%     \midrule
%     \multirow{2}{*}{EgoVLP-Local} 
%                 & EgoVLP-V         & EgoVLP-T       &  \XSolidBrush   &        72.1        & \textbf{91.8}  \\
%                 & EgoVLP-V         & EgoVLP-T       &  \Checkmark     &\textbf{73.8}       &         90.5\\
%     \midrule
%     \multirow{2}{*}{CLIP-ViT-Local} 
%                 & CLIP-ViT-L       & XL-Net         &  \XSolidBrush   &        66.6        & \textbf{90.8}  \\
%                 & CLIP-ViT-L       & XL-Net         &  \Checkmark     &\textbf{67.9}       &         88.9 \\
%     \bottomrule
%   \end{tabular}
%   \caption{Ablations on the effect of the HOI aggregation. HOI denotes hand-object-interaction aggregation module.}
%   \label{tab:3}
% \end{table}

%-------------------------------------------------------------------------
\section{Conclusion}
In this paper, we introduce a Function-Interaction Centric approach with Spatiotemporal Visual-Language Alignment to tackle the AQTC task. Our experimental results showcase the top-1-ranking performance of our proposed method in this competition. Moving forward, we expect the investigation of the context reasoning capabilities of large language models (LLMs) in the domain of AQTC.
%-------------------------------------------------------------------------
\section{Acknowledgement}
This work was partially supported by the National Natural Science Foundation of China (U2013203, 61973106, U1913202, 62103137), the Project of Science Fund for Distinguished Young Scholars of Hunan Province (2021JJ10024); the Project of Science Fund for Young Scholars of Hunan Province(2022JJ40100); the Project of Talent Innovation and Sharing Alliance of Quanzhou City (2021C062L); the Key Research and Development Project of Science and Technology Plan of Hunan Province (2022GK2014).
%-------------------------------------------------------------------------

%%%%%%%%% REFERENCES
{\small
\bibliographystyle{ieee_fullname}
\bibliography{egbib}
}

\end{document}